\documentclass[letterpaper, 10 pt, conference]{ieeeconf}  % Comment this line out if you need a4paper

\IEEEoverridecommandlockouts                              % This command is only needed if 
\usepackage{graphicx} % for pdf, bitmapped graphics files
\usepackage{amsmath} % assumes amsmath package installed
\usepackage{amssymb}  % assumes amsmath package installed
\usepackage{siunitx}
\usepackage{amsfonts}
\usepackage{makecell}
\usepackage{tabularx}
\usepackage{booktabs}
\usepackage{caption}
\usepackage{placeins}

\newcommand{\vct}[1]{\mathbf{#1}}
\renewcommand{\deg}{^{\circ}}
\newcommand{\cruise}{\includegraphics[height=0.8em]{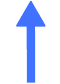}}
\newcommand{\changeleft}{\includegraphics[height=0.8em]{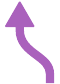}}
\newcommand{\changeright}{\includegraphics[height=0.8em]{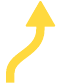}}
\newcommand{\turnleft}{\includegraphics[height=0.8em]{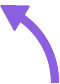}}
\newcommand{\turnright}{\includegraphics[height=0.8em]{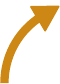}}

\newif\ifanonymized
\anonymizedfalse 

\title{\LARGE \bf
Action Sequence Predictions of Vehicles in Urban Environments\\
using Map and Social Context
}
\ifanonymized 
\author{No authors given at this time.% <-this % stops a space
\thanks{No Institute given at this time.}%
}
\else
\author{Jan-Nico~Zaech$^{1}$\quad Dengxin~Dai$^{1}$\quad Alexander Liniger$^{1}$\quad Luc~Van~Gool$^{1,2}$\\{\tt\small \{zaechj,dai,alex.liniger,vangool\}@vision.ee.ethz.ch}% <-this % stops a space
\thanks{$^{1}$Computer Vision Laboratory, ETH Zurich, Switzerland}%
\thanks{$^{2}$Dept. of Electrical Engineering ESAT, KU Leuven, Belgium}%
}
\fi

\begin{document}

\maketitle
\thispagestyle{empty}
\pagestyle{empty}

%%%%%%%%%%%%%%%%%%%%%%%%%%%%%%%%%%%%%%%%%%%%%%%%%%%%%%%%%%%%%%%%%%%%%%%%%%%%%%%%
\begin{abstract}
This work studies the problem of predicting the sequence of future actions for surround vehicles in real-world driving scenarios. To this aim, we make three main contributions. The first contribution is an automatic method to convert the trajectories recorded in real-world driving scenarios to action sequences with the help of HD maps. The method enables automatic dataset creation for this task from large-scale driving data. Our second contribution lies in applying the method to the well-known traffic agent tracking and prediction dataset Argoverse, resulting in 228,000 action sequences. Additionally, 2,245 action sequences were manually annotated for testing. The third contribution is to propose a novel action sequence prediction method by integrating past positions and velocities of the traffic agents, map information and social context into a single end-to-end trainable neural network. Our experiments prove the merit of the data creation method and the value of the created dataset -- prediction performance improves consistently with the size of the dataset and shows that our action prediction method outperforms comparing models.
\end{abstract}

%%%%%%%%%%%%%%%%%%%%%%%%%%%%%%%%%%%%%%%%%%%%%%%%%%%%%%%%%%%%%%%%%%%%%%%%%%%%%%%%
\FloatBarrier
\section{INTRODUCTION}
\label{sec:intro}
\begin{figure}[t]
    \centering
    \includegraphics[width=1.0\linewidth, trim=150 0 0 45, clip]{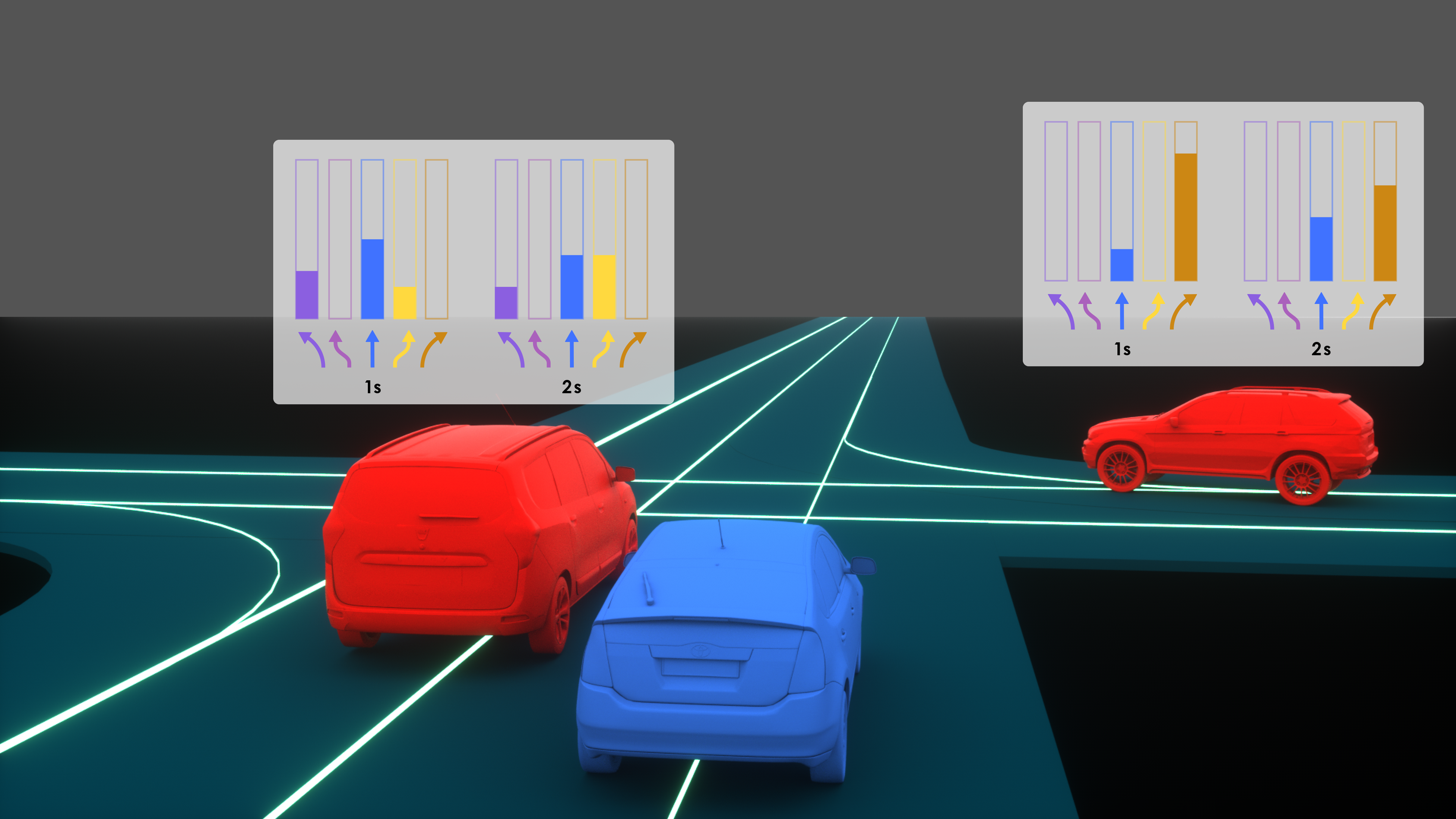}
    \caption{Concept of the proposed method: Based on trajectory and map information, high level action sequences are predicted for other vehicles. This representation is well interpretable and can potentially facilitate downstream planning tasks.}
    \label{fig:intro}
\end{figure}

Autonomous driving is expected to fundamentally change our understanding of mobility and give us safer and more efficient roads. One fundamental building block to achieve this, is the ability to predict future actions of other road users. Only if one is able to accurately predict the potentially multimodal future, collisions can be avoided. However, predicting future actions and trajectories of other road users requires a fundamental understanding of traffic scenarios. This includes understanding the static and dynamic environment, as well as the traffic rules and the unwritten rules that govern how road user interact with each other. 

The most common approach in this research direction is to directly predict trajectories of other vehicles. While this is intuitive, allows for fully automatic data collection with today's test vehicles and yields a good representation for decision making and planning algorithms, it does not consider well that humans learn driving as a sequence of actions and also interpret driving scenarios that way. Furthermore, many traffic rules are defined with high-level representations as driving maneuvers. Thus an autonomous driving system requires emphasis on anticipating high-level actions of surround vehicles to better plan its own actions and to be more interpretable to humans. 

To overcome the aforementioned challenges, we propose to state the problem of predicting traffic agents as the task of predicting action sequences and create a large-scale action prediction dataset based on real driving data. To circumvent the expenditure of annotating a dataset manually, we design an automatic method to convert trajectories of real-world traffic agents into action sequences. We apply the method to the large-scale tracking dataset Argoverse \cite{Argoverse} and compile a new action prediction dataset. The dataset contains 228,000 action sequences and features five distinct driving actions: \emph{cruise} (\cruise, c), \emph{turn left} (\turnleft, tl), \emph{turn right} (\turnright, tr), \emph{lane change left} (\changeleft, ll) and \emph{lane change right} (\changeright, lr) that describe normal vehicle operation. For testing, 2,245 trajectories from the validation set have been manually annotated. The dataset allows us to train and compare models in a quantitative way in terms of their ability to predict a sequence of plausible future actions. To our best knowledge, this is the largest dataset for action prediction of traffic agents derived from publicly available data.

We further propose an end-to-end trainable neural network that uses the past positions and velocities of the traffic agents, the map information and the social context to predict the future actions of the traffic agents. To fully leverage the power of convolutional neural networks (CNNs), we encode these information into a rendered image with multiple channels and use 2D and 3D CNN modules for prediction. Because the future actions are often uncertain and multimodal, out model outputs a probabilistic distribution of all plausible actions which can facilitate downstream planning.  Fig.\ref{fig:intro} showcases the concept of our action prediction approach. 

\section{Related Work}\label{sec:related_work}
Behaviour prediction formulated as trajectory forecasting for humans as well as vehicles has been extensively studied.

The line of research closest to ours focuses on predicting high level intention of traffic agents. One group of work in driver action prediction concentrates on the ego vehicle where rich information about the state is available. Morris et al. \cite{morris_lane_2011} use rich sensor data including radar, lane marking detection and a head tracking camera to predict lane changes in a highway driving scenario. Jain et al. \cite{jain_car_2015} extend this approach to a larger set of maneuvers and base their method on a video of the driver, maps, vehicle dynamics and outside view in more diverse environments. In \cite{hecker_endtoend_2018} surround video and map renderings are used to predict yaw and acceleration in an end-to-end framework.
\cite{schlechtriemen_lane_2014} forecast lane-changes of other traffic agents in highway scenarios and analyze the challenge of heavily unbalanced data in this context. In more challenging urban environments, \cite{khosroshahi_surround_2016} classify driving actions at structured four way intersections with an LSTM.

Early approaches to trajectories prediction, combine maneuver recognition with parametric motion models for each maneuver. Laugier et al. \cite{laugier_probabilistic_2011} use a Hidden Markov Model (HMM) with access to high level information such as distance to lane borders, signaling light status or proximity to an intersection to recognize behaviour of traffic agents. Trajectories are then sampled from a Gaussian process and used for evaluating collision risks. The same task is approached in \cite{schreier_bayesian_2014} by detecting maneuvers with a Bayesian network. Houenou et al. \cite{houenou_vehicle_2013} propose a heuristic maneuver detection module with full environment knowledge including each vehicle's acceleration and yaw angle, together with an analytic description of trajectory sets. By using a variational Gaussian Mixture Model, Deo et al. \cite{deo_how_2018, deo_multi-modal_2018} implement probabilistic trajectory prediction for highway scenarios in the combined maneuver and trajectory prediction framework. Recently, \cite{casas_intentnet_2018} used a neural network based approach to combine intention and trajectory prediction with dynamic HD maps and LIDAR information.

A second group of trajectory prediction algorithms directly approaches the task without intermediate state representations. Lee et al. \cite{lee_desire_2017} propose an end-to-end trainable recurrent neural network structure that includes scene context and samples multiple trajectories to capture the multi-modal nature of trajectory prediction. In \cite{Hong_2019_CVPR}, a wide range of output representations are evaluated in combination with a fully convolutional encoder structure. By defining a graph structure, \cite{li_grip_2019-1} explicitly models the relation between multiple traffic agents and uses an LSTM-based encoder-decoder to predict trajectories. Closely related to trajectory prediction, researchers at Waymo \cite{bansal_chauffeurnet_2018} learn a driving policy using rich maps and employ data augmentation to train robust models.

In the context of pedestrian prediction and tracking, a central challenge is modeling of interactions. Pellegrini et al. \cite{pellegrini_youll_2009} show that social interactions and scene knowledge can boost tracking performance. \cite{alahi_social_2016, gupta_social_2018} predict trajectories based on past ego and social trajectory observations, while matching not only trajectories but also distributions. Sadeghian et al. \cite{Sadeghian_2019_CVPR} add world context using a top down view and add attention to decide what part of the world and social aspect are important.

\section{Method}
First, our method introduces an interpretable representation of traffic agents by forecasting high level action sequences of a single traffic agent. Second, our method is completely learning based and uses a sequence of map, agent, as well as social information (other traffic agents) to predict the agent action sequence. This is done using a fully convolutional neural network with two-dimensional convolutions for computing spatial features and two three-dimensional layers for late and early fusion of temporal features. The networks head is a fully connected layer that predicts the full action sequence of the traffic agent at once with a single forward pass through the network. Finally, to make this method applicable for large scale datasets, we introduce an automatic approach to generate action sequences from X-Y trajectory data and map information. Note that our method is \emph{entity-centric} and only predicts the action sequence of one agent, however we consider other traffic agents during prediction.

\subsection{Network Architecture}
We use a VGG-inspired architecture for action sequence prediction. Like \cite{Hong_2019_CVPR}, early and late fusion of temporal features is performed with three-dimensional convolutions. To avoid overfitting on the training data, we use a smaller model compared to some fully convolutional trajectory prediction approaches \cite{Hong_2019_CVPR}.
\begin{figure}[tb]
    \vspace{4pt}
    \centering
    \includegraphics[width=0.9\linewidth, trim=45 120 495 150, clip]{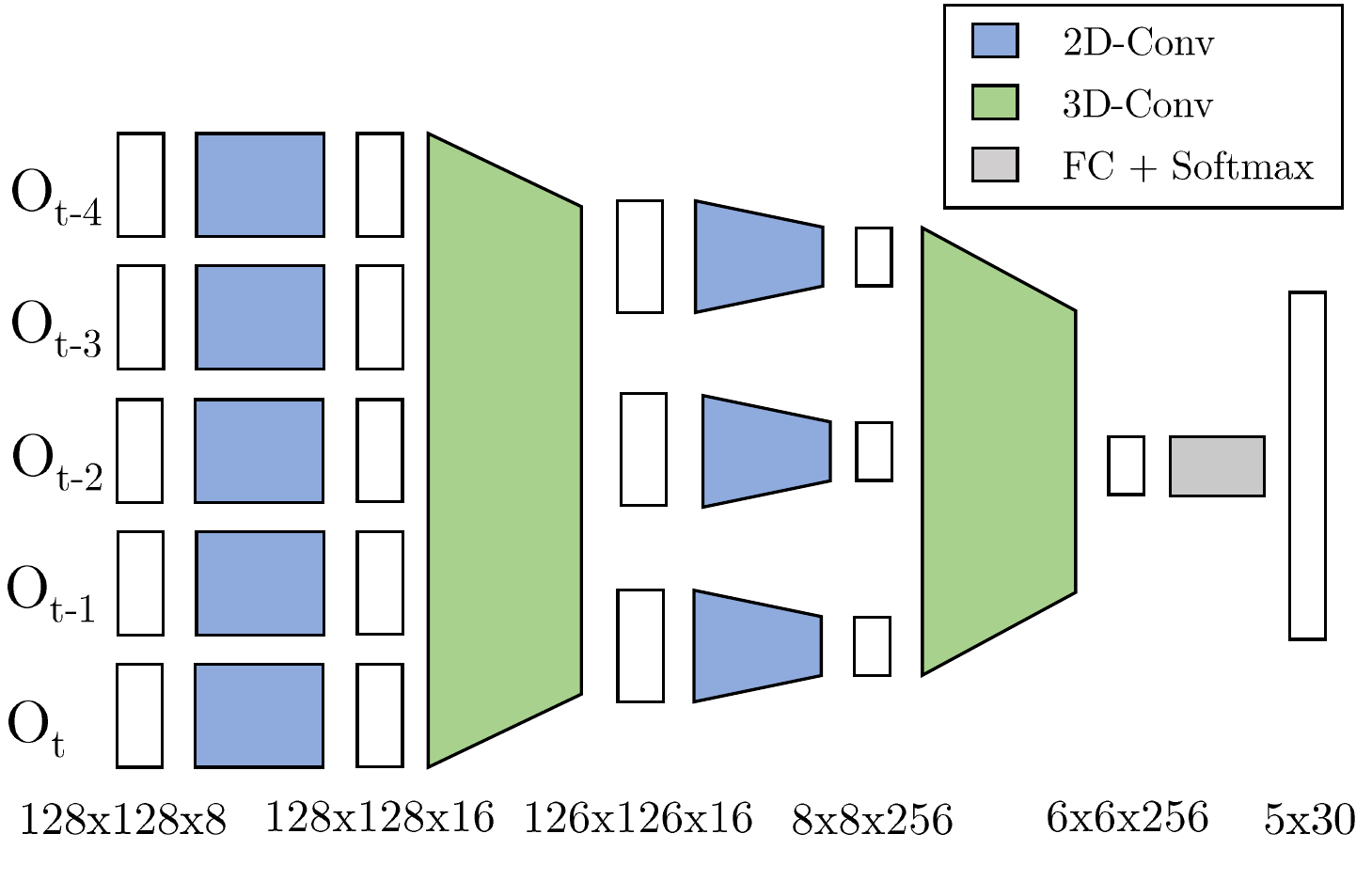}
    \caption{VGG inspired network architecture, with 2D convolutions to extract spatial features and 3D convolutions for early and late temporal fusion.}
    \label{fig:network_architecture}
\end{figure}

\subsection{Input and Output Representation}
The input to our neural network consists of three parts, target agent, map and social information. All the information comes as a tensors with spatial dimension $128\times128$. We consider this combined information as our observation at time $t$, which we denote as $\vct{O_t}$. More precisely, $\vct{O_t}$ is given as, 
\begin{equation} \label{eq:obs}
\vct{O_t} = \left(\vct{m}, \vct{1}_{target}, \vct{v}_{target}, \vct{1}_{other}, \vct{v}_{other}\right)_t \,,
\end{equation}
where, all components are rendered frames, spanning \SI{50x50}{\metre}. All frames are centered at the target agent's last observed position and rotated towards its driving direction. Each observation $\vct{O_t}$ contains one layer for the rendered lane centerlines $\vct{m}$ which stays fixed for all time steps. As the map only consists of centerline information, it can easily be captured with current industry grade perception systems and is present in most rich maps. Still, it naturally extends to more extensive information that might be available from manual annotations or more advanced data acquisition systems. Additionally, in $\vct{O_t}$ the target agent is represented by three layers: $\vct{1}_{target,t}$, which is the indicator function representing the target agent's position by a one-hot encoded layer and two layers $\vct{v}_{target,t}$ representing the current velocity in global coordinates. The other agents are represented the same way, but with all agents jointly rendered into the three available channels. 

Finally, the input to the network are the last five observation,
\begin{equation}
\vct{O} = \left(\vct{O}_{t_{-4}}, \vct{O}_{t_{-3}}, \vct{O}_{t_{-2}}, \vct{O}_{t_{-1}}, \vct{O}_{t_{0}} \right) \,,
\end{equation}
where the observations are spread over the last two seconds, with $t \in (\SI{-2}{\second}, \SI{-1.5}{\second}, \SI{-1}{\second}, \SI{-0.5}{\second}, \SI{0}{\second})$.

To improve the performance and robustness of our method we also perform data augmentation. More precisely, we rotate the network input randomly by $\theta$, which is uniformly sampled from the range $-5\deg \leq \theta \leq 5\deg$. To perform the augmentation efficiently and without artifacts, all input data is stored in parametric form and rendered on-the-fly.

The output of the network is an action sequence for the next \SI{3}{\second}, resulting in a prediction horizon of 30 steps with a sampling time of \SI{100}{\milli\second}. The output of the network model at each time step is an independent probability distribution of the 5 action classes. Note that the temporal relation and implicit dependence needs to be learned by the model from the training data.

\subsection{Probabilistic Action Predictions}
In contrast to trajectory forecasting methods, our approach directly returns probabilistic predictions, without any requirement for sampling multiple forecasts \cite{lee_desire_2017} or defining spatially discretized grid-maps, as done in \cite{kim_probabilistic_2018, Hong_2019_CVPR}. Furthermore, this also allows for transparent performance measures since action classes can easily be interpreted by humans. The performance evaluation of trajectory prediction methods on the other hand normally uses averaged displacements errors. Even thought this seems to be an transparent evaluation, due to inbalanced datasets where following the current lane (our cruise action) is heavily over-represented, getting better displacement errors not necessarily implies that the method is better at forecasting the important corner cases. Note that the imbalance in the datasets can be massive, e.g. our dataset consists of $80\%$ cruise trajectory, however, there are other traffic environments where cruise trajectories can make up as much as $99\%$ of the dataset \cite{schlechtriemen_lane_2014}. This imbalance also explains that often simple constant velocity forecasting methods are competitive for short prediction horizons \cite{Hong_2019_CVPR}, even thought they lack any understanding of the traffic scenario.

By predicting action sequences, our method does not solve the imbalance in the dataset. However, our evaluation can be balanced easily and we can focus on complex scenarios that require to understand the traffic scene.

\section{Dataset}
Many large scale datasets for visual tasks in autonomous driving such as image segmentation, object detection or depth estimation have been released in recent years and fueled the development of corresponding learning based methods \cite{Argoverse,geiger_vision_2013,kesten_lyft_2019,nuscenes2019}. In contrast to this, modeling and prediction of human decision making in driving scenarios is heavily underrepresented in terms of published methods. One reason for this definitely is the lack of data. Public datasets, while being suitable for vision tasks, fall short when approaching the task of modeling human driving behaviour in complex environments. Furthermore, most datasets directly intended for this purpose remain private \cite{casas_intentnet_2018,Hong_2019_CVPR}.

\subsection{Argoverse}
An exception is the Argoverse Trajectory Forecasting dataset \cite{Argoverse}, which contains approximately 325,000 automatically detected trajectories in an urban environment. Out of them, 245,000 cover 5s segments that can by used for our approach\footnote{80,000 trajectories are in the test set and only show 2s segments.}. The dataset further provides basic semantic map information, including lane centerlines and the drivable area. In our work we augment this dataset by automatically annotating high-level actions such as lane changes and turns that are interpretable by humans and have the potential to boost low level tasks like trajectory forecasting \cite{deo_multi-modal_2018}. Finally, we also automatically extract velocity information of the agents, which further helps our prediction model.

\subsection{Map Information}
As shown and discussed in Section \ref{sec:related_work}, using HD-map information can be fundamental for traffic agent forecasting. However, in this paper we show that HD-maps can also be used to automatically annotate data, or in our case generate high-level action sequences from trajectories. This avoids, time and labor intensive manual labeling, and at approximately 245,000 trajectories in Argoverse, which corresponds to roughly 7.5M action annotations, this is the only viable approach. 

One the one hand, relying on HD-maps for the action sequence generation, is in our opinion not restrictive, since maps are regarded essential for safe operation of autonomous vehicles. On the other had, for large scale datasets, annotating a semantic map with lane centerlines, which remains constant over time, scales favorably compared to annotating every time step of every agent recorded during driving. Furthermore, as current lane detection algorithms show an impressive performance in a wide range of practical scenarios, automatic extraction of local map information could be feasible to further automate the labeling process.

\begin{figure}[tb]
    \centering
    \includegraphics[width=1.0\linewidth, trim=130 110 380 80, clip]{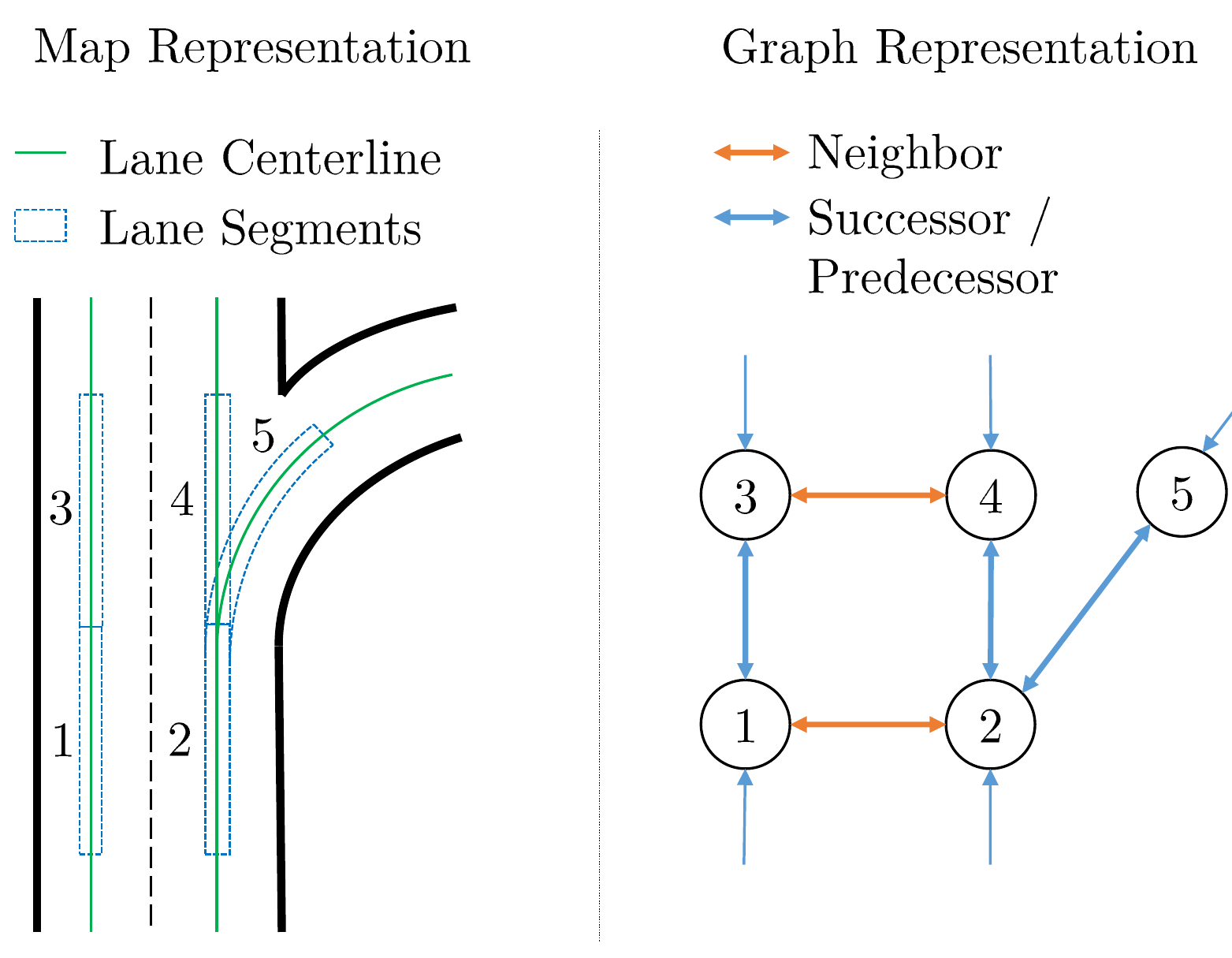}
    \caption{Map represented as rendered centerlines and as grap-map. Each centerline corresponds to a node in the graph-map which are connected by edges labeling their semantic relation.}
    \label{fig:map_representation}
\end{figure}

For the task of action sequence annotation, the semantic HD-map available in \cite{Argoverse}, can be represented as a graph where each node corresponds to a short sequence of line segments and edges represent the relation of between the line segments, which is visualized in Fig. \ref{fig:map_representation}. Edges can have the labels successor, predecessor, and neighbor lane-segment. Nodes contain the geometric properties describing the lane segments and semantic information if the segments describe a turn (either left or right) or a lane driving straight forward.

\subsection{Automatic Action Labeling}\label{sec:action_labeling}
Action extraction from trajectories is performed in a three-stage pipeline described in the following paragraphs.

\paragraph{Trajectory Smoothing}
As the trajectories are generated from automatically detected objects, all samples are noisy and need to be filtered in the first stage. For this purpose we use a bidirectional Kalman filter with a standard constant acceleration model, where the jerk is modelled as a noise input and we use the agent's noisy position as the measurement. Note that this filter does not only help to smooth the agent's position but also estimates the velocity of the agent.

\paragraph{Lane Assignment}
In the second stage of the annotation pipeline, each sample from the trajectory is assigned to a node, which represents a lane-segment in the graph-map as shown in Fig. \ref{fig:map_representation}. Following the temporal order of samples, each trajectory induces a sequence of nodes that are visited. If the sequence of nodes follows a valid path through the graph, i.e. the path only contains transitions between nodes that are connected by an edge, actions can be extracted from the graph. The property that only a valid path allows for the extraction of an action sequence has major impact on designing the node assignment algorithm. Using a purely geometric approach, where each sample is assigned to its nearest neighbor in an arbitrary measure, could lead to a large fraction of invalid trajectories. We thus propose to use a joint geometric and semantic lane assignment algorithm based on the Viterbi algorithm.

One can interpret a trajectory as the observation from a Hidden Markov Model (HMM) defined by the graph-map together with actions as latent variables. The graph-map induces a HMM, where the lane segment used by the agent at time step $t$ corresponds to the hidden state $X_t$ and the position $Y_t \in \mathbb{R}^2$ is the symbol emitted by the state. The emission probability 
\begin{equation}
    P(Y_t | X_t = x_t)
\end{equation}
captures the driver's behaviour of not perfectly following the centerline, uncertainties in the map and measurement noise. Using this viewpoint, the actions taken by the agent define the sequence of hidden states, or inversely: inferring the most likely sequence of hidden states from the observed trajectory is equivalent to finding the underlying action sequence. This task can be solved with the Viterbi algorithm.

To reduce computational complexity, only lane segments that are closer than a threshold of \SI{5}{\metre} to the observed trajectory are included as hidden states in the HMM. Prior knowledge about the driving behaviour can be encoded in the transition matrix $\vct{A}$ between lane segments, which we populate using the edges of the graph-map that describe the relation between lane segments. For the possible transitions to a successor, predecessor or neighbor lane, values of $1.0$, $0.5$ and $0.3$ are assigned respectively. For transitions between not connected lanes, a skew variable $\alpha$ is introduced. Setting $\alpha = 0$, yields state sequences that can be annotated in any case, given the map information is correct. This also includes trajectories that actually cannot be modelled by the 5 action classes used in this work, such as U-turns. Thus, we set $\alpha = 0.001$ which allows for transitions between unconnected lane segments, preventing the annotation of an action sequence for cases that cannot be annotated with the given actions. It is important to note that for the transition matrix $\vct{A}$ the sum of entries from a state does not necessarily sum up to 1 and thus, does not represent a traditional transition probability matrix, but rather a matrix of negative transition costs. We use this representation to not penalize lane-segments that have multiple lanes connected to them in the graph-map, such that trajectories going through them do not come at a higher cost.

\paragraph{Action Extraction}
With the most likely sequence of lane-segments determined by the Viterbi algorithm, actions can be extracted from the map information with a rule based system. Lane changes are labeled for transitions between lane segments marked as neighbors\footnote{Corner cases such as the direct change to the neighbor of a successor lane need to be modeled and handled adequately.} and turns can directly be labeled from the turn annotation in the lane segment.

All actions are handled as non-singular events, i.g. the lane change state is annotated for all time steps between leaving the previous lane and stabilizing on the new lane. To enable this, the smoothed trajectory together with the lane geometry is used to extract the start and end point of all maneuvers.

\subsection{Test Data and Ordered Action Sequences}
For testing, $2245$ trajectories from the validation set have been manually annotated. However, the annotator did not generate temporal action sequences, but what we call ordered action sequences. Whereas action sequences, as produced by our automatic labeling, contain an action for each time step, ordered action sequences only contain the order of actions. To make this clear, let us consider an simple left lane change example where the prediction horizon is six time steps. In our example the the action sequence is given by $\vct{a}_s = ($c,c,ll,ll,c,c$)$. This action sequence would correspond to the following order action sequence $\vct{a}_{os} = ($c,ll,c$)$, thus the exact temporal location of the lane change is lost, however, the gist of the maneuver is captured. 

Annotating, ordered action sequences is also significantly simpler and less error prune, compared to temporal action sequences. Thus, for our $2245$ test trajectories, we have hand annotated ordered action sequences for the \SI{3}{\second} prediction horizon. Note that to avoid bias, the annotator had no access the automatically generated labels.

\subsection{Dataset Statistics}
While the trajectories present in the Argoverse dataset are already filtered to show challenging behaviour, the extracted action data is still unbalanced, with the majority of the samples representing the cruise class as shown in Table \ref{table:dataset_distribution}. While this mostly stems from the lower probability of encountering an active maneuver compared to just following a lane in cruise, it also reflects the fact that a turn or lane-change is most of the times followed by the traffic agent stabilizing in the cruise state. 

Whereas generating the dataset statistics for our automatically generated action sequences in the training and validation set is based on the number of occurrences, the statistics for the ordered action sequences used an adapted method. State proportions are estimated by counting states with the inverse number of total states annotated for the corresponding sequence, e.g. a sequence only annotated as cruise counts $100\%$ towards cruise, while a sequence consisting of cruise and a following lane change counts with $50\%$ towards both classes. In total, the state distribution for the train, validation, and test set is as shown in Table \ref{table:dataset_distribution}. 

\begin{table}[]
\vspace{3pt}
\begin{center}
\small
\begin{tabular}{c|c|ccccc}
\multicolumn{1}{c }{}   &
\multicolumn{1}{c }{}   &
\multicolumn{1}{c }{} &
\multicolumn{2}{c }{Turn}   &
\multicolumn{2}{c}{Change}   \\
Split
&Total
&Cruise
&left
&right
&left
&right\\

\hline
train
&191,841
&84.5\%
&7.6\%
&4.0\%
&2.1\%
&1.9\%\\
%\hline
val
&36,471
&87.5\%
&4.0\%
&3.2\%
&2.9\%
&2.4\%\\
manual
&2,245
&87.2\%
&4.1\%
&2.7\%
&3.5\%
&2.6\%\\
\end{tabular}
\end{center}
\caption{Label distribution of the annotated data.}
\label{table:dataset_distribution}
\end{table}

\section{Experiments}

\subsection{Training}
Network parameters are optimized with the Adam optimizer and an initial learning rate of $10^{-4}$. A step decay schedule with a stepsize of $10$~epochs and a factor of $0.5$ is used for adapting the learning rate. Dropout of~$0.5$ is used to reduce overfitting. All variations of the network are trained for 50 epochs.

To compensate for the heavily imbalanced dataset, weighted random sampling is used for dataloading. Weights are assigned based on the presence of actions anywhere in the prediction horizon: if a turn is present, the sample gets weighted with a factor of 3, if a lane-change is present, the weighting factor is set to 10. This does not fully compensate for the imbalance, but performed better than strictly weighting samples by their inverse probability.

Note that the positions and velocities of all traffic agents used in the input representation \eqref{eq:obs} are extracted from the noisy trajectory data with a bidirectional Kalman-filter, as described in Section \ref{sec:action_labeling}. However, the filter is only applied to the observed values, to ensure causality.

\subsection{Baseline Model}
By proposing a new formulation of the traffic agent prediction task, no direct comparison to other methods would be possible. However, trajectory forecasting methods are intended to predict a trajectory that represents the future actions of the agent. Therefore, it is possible to adapt a method originally designed for trajectory prediction to the new task. For our evaluation, a k-Nearest-Neighbor (k-NN) based method that was evaluated in \cite{Argoverse}\footnote{arXiv:1911.02620 [v1] 6 Nov 2019} and outperformed all their tested deep learning models for multimodal prediction is adapted to our problem statement and used as a baseline-model for comparison. With the nearest neighbor in the training set, the automatically extracted action sequences can be used as a prediction for the task at hand. Given the predictions from k-NN, the class labels can be averaged at each time step, resulting in a prediction that matches the probabilistic form of our proposed approach. We set $k \in \{9,50,100\}$ to compare the influence of sampling multiple trajectories and select the best model for further comparison.

\subsection{Evaluation Approach}
\subsubsection{Direct Evaluation of Predictions}
Interpreting every time step as an independent classification problem, allows for the direct evaluation of the networks performance. As the data is heavily imbalanced, classification accuracy on the whole dataset does not provide sufficient insight. Therefore, we measure performance by representing every action class as a binary classification problem, which allows the calculation of the average precision (AP) score. AP for each class and their unweighted average denoted as mean AP across all five action classes are used as a performance measure. This direct evaluation approach jointly measures the performance of predicting the right action classes together with the performance of predicting the right time step for a transition between two actions.
\begin{table}[]
\vspace{3pt}
\begin{center}
\small
\begin{tabular}{ c | c | c c c c c}
\multicolumn{1}{c }{}   &
\multicolumn{1}{c }{}   &
\multicolumn{1}{c }{}   &
\multicolumn{2}{c }{Turn} &
\multicolumn{2}{c }{Change}\\
Method
&Mean
&Cruise
&left
&right
&left
&right\\
\hline
random
&20.0
&87.5
&4.0
&3.2
&2.9
&2.4\\
k-NN (9) \cite{Argoverse}
&32.5
&93.5
&32.9
&26.3
&5.4
&4.2\\
k-NN (50) \cite{Argoverse}
&36.9
&95.0
&39.2
&34.7
&8.2
&7.1\\
%\hline
k-NN (100) \cite{Argoverse}
&37.4
&95.3
&39.7
&35.4
&8.6
&8.0\\
%\hline
ours
&\bf{61.4}
&\bf{97.8}
&\bf{67.9}
&\bf{68.4}
&\bf{33.5}
&\bf{39.4}\\
\end{tabular}
\end{center}
\caption{AP scores of the investigated methods on the automatically annotated validation set. Random sampling performance corresponds to dataset proportion for each class. All scores in \%.}
\label{table:direct_performance}
\end{table}

\subsubsection{$N$-Most Likely Ordered Action Sequences}
In addition to directly measuring the model's precision on the temporal action predictions, we evaluate our method by extracting the $N$-most likely ordered sequences of actions, i.e. sequences that are only defined by the order of actions, not their exact length or transition times. 

The extraction of the $N$-most likely ordered action sequences, is sensible as the number of different actions in the \SI{3}{\second} prediction horizon is small. For our test set only $10$ samples, corresponding to $0.45\%$ of the manual annotations, where labeled as as sequence of more than two actions. Therefore, for extracting ordered action sequences from the temporal action sequence predictions, we only consider sequences with at most two actions, which makes the approach tractable. 

When extracting ordered action sequences, the model of independent actions used during training needs to be taken into account. While this assumption is a good model for the probability of an agent performing action $a_t$ at time step $t$, it is not suitable to approximate the probability of observing a full action sequence $\vct{a}_s = (a_0, ... a_T)$. By just using the product of the predicted probabilities
\begin{equation}
    p_{ind} = \prod_t\hat{p}_{t}^{a_t},
\end{equation}
as an estimate for the sequence probability, the high temporal correlation of actions is neglected. We thus model the probability of a sequence with the two actions $(a_{b1},a_{b2})$ by
\begin{equation}
    p_b(a_{b1},a_{b2},t_s) = \min(\hat{p}_{t_0}^{a_{b1}}, ..., \hat{p}_{t_{s}}^{a_{b1}} )\min(\hat{p}_{t_{s}+1}^{a_{b2}}, ..., \hat{p}_{T}^{a_{b2}}),
\end{equation}
where the transition happens after time step $t_s$ leading to the blocks $t_0 \leq t \leq t_s$ of $a_{b1}$ and $t_s < t \leq T$ of $a_{b2}$. Given that within a block no transition may happen, the transition probabilities between identical actions are $1$. Therefore, the total probability of the first block can be modelled as the lowest predicted probability for $a_{b1}$ within $t_0 \leq t \leq t_s$
\begin{equation}
    \min(\hat{p}_{t_0}^{a_{b1}}, ..., \hat{p}_{t_{s}}^{a_{b1}} ).
\end{equation}
Note that the same holds for $a_{b2}$ and the second block. Under the assumption that rare combinations, such as a left turn directly followed by a right turn, are assigned low probabilities by the network, the transition probabilities between the blocks are modeled as being equal for all action pairs $(a_{b1},a_{b2})$.

The problem of finding the most likely ordered sequence of two actions can then be defined as finding $(a_{b1}, a_{b2}, t_s)$ that maximize the product of the two block probabilities
\begin{equation}
    p_{b,opt} = \max_{t_{s}, a_{b1}, a_{b2}} p_b(a_{b1}, a_{b2}, t_s).
\end{equation}
By repeatedly searching for the most likely sequence while suppressing already extracted action pairs $(a_{b1},a_{b2})$, the $N$-most likely ordered sequences can be extracted. We report total and per sequence top-$N$ accuracy with $N \in \{1,2,3\}$ in Table \ref{table:n_most_likely_performance}. A sequence is rated as being detected if the groundtruth ordered state sequence is one of the top-$N$ predictions. Trajectories that are annotated with sequences of length $>2$ are always treated as being detected incorrectly.

\begin{table}[]
\vspace{3pt}
\begin{center}
\small
\begin{tabular}{ l l | r  r  r | r  r  r}
\multicolumn{2}{c }{}   &
\multicolumn{3}{c }{Ours} &
\multicolumn{3}{c}{k-NN (100) \cite{Argoverse}} \\
\multicolumn{2}{c }{Sequence}
&Top1
&Top2
&Top3
&Top1
&Top2
&Top3\\
\hline
\multicolumn{2}{l |}{total}			&	82.5	&	89.8	&	93.0	&	81.2	&	86.4	&	88.6	\\
c 	&	\cruise	&	94.1	&	96.2	&	97.9	&	99.6	&	99.8	&	99.9	\\
ll, c 	&	\changeleft\cruise	&	65.7	&	85.7	&	95.7	&	1.4	&	27.1	&	41.4	\\
tl 	&	\turnleft	&	63.3	&	69.4	&	73.5	&	14.3	&	18.4	&	26.5	\\
lr, c 	&	\changeright\cruise	&	44.7	&	72.3	&	83.0	&	0.0	&	36.2	&	53.2	\\
tl, c 	&	\turnleft\cruise	&	25.0	&	65.0	&	87.5	&	32.5	&	95.0	&	97.5	\\
tr, c 	&	\turnright\cruise	&	42.5	&	75.0	&	80.0	&	17.5	&	70.0	&	87.5	\\
c, tl 	&	\cruise\turnleft	&	12.1	&	69.7	&	75.8	&	0.0	&	42.4	&	69.7	\\
c, tr 	&	\cruise\turnright	&	24.2	&	72.7	&	78.8	&	0.0	&	36.4	&	48.5	\\
c, ll 	&	\cruise\changeleft	&	4.3	&	52.2	&	73.9	&	0.0	&	17.4	&	26.1	\\
ll 	&	\changeleft	&	23.8	&	38.1	&	42.9	&	0.0	&	0.0	&	0.0	\\
lr 	&	\changeright	&	11.1	&	22.2	&	22.2	&	0.0	&	0.0	&	0.0	\\
tr 	&	\turnright	&	35.3	&	47.1	&	58.8	&	0.0	&	0.0	&	5.9	\\
c, lr 	&	\cruise\changeright	&	6.3	&	75.0	&	75.0	&	0.0	&	6.3	&	6.3	\\
tr, ll 	&	\turnright\changeleft	&	14.3	&	21.4	&	42.9	&	0.0	&	0.0	&	0.0	\\
tl, lr 	&	\turnleft\changeright	&	40.0	&	50.0	&	60.0	&	0.0	&	0.0	&	10.0	

\end{tabular}
\end{center}
\caption{Evaluation of the Top-$N$ ordered sequence predictions. The shown sequences are ordered by descending number and at least have 10 samples. All scores are accuracies in \%.}
\label{table:n_most_likely_performance}
\end{table}

\section{Results}

\subsection{Direct Evaluation}
Performance metrics for the direct evaluation of predictions for the k-NN baseline \cite{Argoverse} and our proposed method are shown in Table \ref{table:direct_performance}. The results affirm that increasing the number of sampled nearest neighbors by one magnitude (k=100) compared to the usual setting can improve its performance. Across all methods, the AP for the cruise state is on a high level, followed by turn actions. Lane changes are harder to predict with AP values at a much lower level. Also, the largest relative improvement of the proposed method compared to the baseline implementation can be seen for the two lane change action classes.

This may be explained by the dataset statistics together with shape of lane change trajectories. While lane changes as well as turns are underrepresented in the training data, turns have a much more distinct trajectory shape and thus can be detected more easily, even without modeling them explicitly.

\subsection{$N$-Most Likely Ordered Sequences}
Results of evaluating our proposed as well as the best performing baseline method on the manually annotated ordered sequences are provided in Table \ref{table:n_most_likely_performance}. In contrast to the baseline model, where the cruise action class is overrepresented, our proposed method predicts a more diverse set of action sequences. While a relatively high Top1-accuracy is observed for sequences that require detecting a maneuver, e.g. (tl), (ll, c), (lr, c), sequences that involve the prediction of maneuvers such as (c, ll) or (c, lr) show a high improvement in the Top2-accuracy. The observation that some action classes heavily improve in Top2-accuracy is in line with the commonly accepted assumption that agent prediction needs to be multimodal to account for the uncertainty of the future.

\begin{figure}[tb]
    \centering
    \includegraphics[width=0.9\linewidth, trim=17 17 19.5 19, clip]{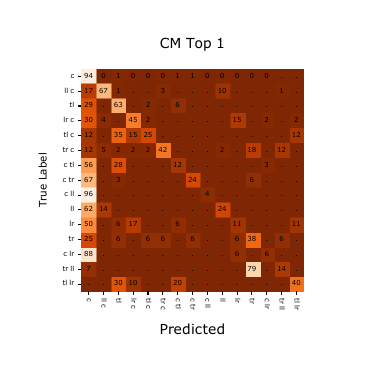}
    \caption{Confusion matrix for top-1 ordered sequence predictions with the x-axis showing predicted classes and y-axis showing groundtruth classes. All numbers in \%, normalized that rows sum to 100.}
    \label{fig:cm_top1}
\end{figure}
\begin{figure}[tb]
    \centering
    \includegraphics[width=0.9\linewidth, trim=17 17 19.5 19, clip]{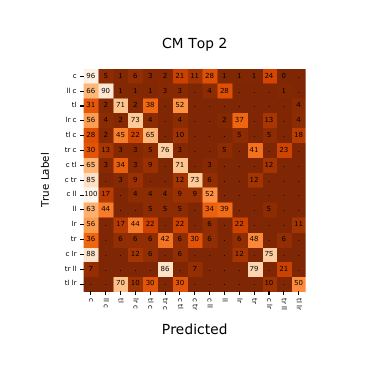}
    \caption{Confusion matrix for top-2 ordered sequence predictions with the x-axis showing predicted classes and y-axis showing groundtruth classes. All numbers in \%, normalized that rows sum to 200.}
    \label{fig:cm_top2}
\end{figure}

Fig. \ref{fig:cm_top1} and \ref{fig:cm_top2} show the confusion matrix for the top-1 and top-2 sequence prediction, with sequences ordered according to their frequency in the dataset. For top-1 prediction, the classification errors mostly stem from assigning sequences with an incorrect second action, e.g. (tl) instead of (tl, c), which confirms that prediction is a much harder task than classification. Still, the top-2 predictions allow for correcting for many of these errors, resulting in substantially higher values on the diagonal.

\subsection{Ablation Study}
An ablation study is conducted to evaluate the impact of the separate modules. Ablated methods are compared using the mean AP on the validation dataset, with results shown in Table \ref{table:network_ablation}. To investigate the influence of input representation the input tensor is grouped into following modules: The target agent information (Target) includes positions and velocities of the target agent $\left(\vct{1}_{target}, \vct{v}_{target}\right)$. Social context (Social) comprises of the positions and velocities of the other agents $\left(\vct{1}_{other}, \vct{v}_{other}\right)$. Finally, map information (Map) contains the layer representing the lane centerlines~$\vct{m}$.

Besides ablating the input representation, we show that traffic agent action prediction can profit from the presented data augmentation approach of rotating the complete contextual information by a small random angle.

\begin{table}[]
\begin{center}
\small
\begin{tabular}{ c  c  c | c | c }
\multicolumn{3}{c }{Input representation}   &
\multicolumn{1}{c }{}   & 
\multicolumn{1}{c}{}\\
Target & Social & Map & Augmentation & mean AP\\
\hline
\checkmark & & & \checkmark & 0.375\\
\checkmark & \checkmark & & \checkmark & 0.490\\
\checkmark& & \checkmark & \checkmark & 0.602\\
\checkmark& \checkmark& \checkmark & & 0.590\\
\checkmark& \checkmark& \checkmark & \checkmark  & \textbf{0.614}\\
\end{tabular}
\end{center}
\caption{Ablation study on the action prediction network.}
\label{table:network_ablation}
\end{table}

The results confirm the usefulness of providing all three sources of information jointly, with the map having the biggest impact. Interestingly, adding social information to a representation that does not contain a map, has substantially higher impact than adding it to a representation that does contain a map. This indicates that there is redundant information available in the map and social context the network is able to use.

\subsection{Dataset Scale study}
The second ablation study investigates the relevance of the dataset size for the action prediction task while using our full model. Thus, the network is trained with three subsets of our data that reflect 50\%, 10\% and 5\% of the full set of action sequences. The dataset reduction is implemented by skipping the corresponding number of samples, to avoid changing the dataset statistics. Epoch length and sample weights are kept constant, such that the training parameters are stable. The results of the ablation study, shown in Table \ref{table:dataset_ablation}, show a large difference in performance between the different scales, indicating that the amount of data has major impact on the prediction performance. Furthermore, the considerable improvement from 50\% of the data to the full dataset allows for the conclusion that the performance did not saturate at the scale present in the Argoverse dataset and larger amounts of data could benefit the community.
\begin{table}[]
\vspace{3pt}
\begin{center}
\small
\begin{tabular}{ c | c }
Dataset proportion & mean AP\\
\hline
100\% & \bf{0.614}\\
50\% & 0.571\\
10\% & 0.522\\
5\% & 0.493\\
\end{tabular}
\end{center}
\caption{Ablation study on the action prediction dataset.}
\label{table:dataset_ablation}
\end{table}
\section{Conclusion and Future Work}

In this paper we investigated the task of predicting high-level actions of vehicles in urban environments. To make this task viable, we proposed an algorithm to automatically extract action sequences using HD-maps, from the public Argoverse \cite{Argoverse} dataset. Furthermore, we proposed an action prediction network, that predicts the future actions sequence considering agent, map, and social information. The network is completely based in rendered images and can be trained in an end-to-end fashion using the automatically generated large scale action sequence dataset. We showed that our action prediction model, together with our dataset, can significantly outperform existing methods that are adapted from trajectory prediction. This additionally shows that action prediction is not solved by trajectory prediction. Thus, in our future work we will investigate how action prediction and trajectory prediction can be combined to get the best of both worlds.

\addtolength{\textheight}{0cm}   % This command serves to balance the column lengths

%%%%%%%%%%%%%%%%%%%%%%%%%%%%%%%%%%%%%%%%%%%%%%%%%%%%%%%%%%%%%%%%%%%%%%%%%%%%%%%%

%%%%%%%%%%%%%%%%%%%%%%%%%%%%%%%%%%%%%%%%%%%%%%%%%%%%%%%%%%%%%%%%%%%%%%%%%%%%%%%%

%%%%%%%%%%%%%%%%%%%%%%%%%%%%%%%%%%%%%%%%%%%%%%%%%%%%%%%%%%%%%%%%%%%%%%%%%%%%%%%%
\vspace{1mm}
\noindent
\ifanonymized 
\textbf{Acknowledgement}: No acknowledgment given at this time.
\else
\textbf{Acknowledgement}: The work is supported by Toyota Motor Europe via the research project TRACE-Z\"urich.
\fi

%%%%%%%%%%%%%%%%%%%%%%%%%%%%%%%%%%%%%%%%%%%%%%%%%%%%%%%%%%%%%%%%%%%%%%%%%%%%%%%%
%\bibliographystyle{IEEEtran}
%\bibliography{references}

\end{document}